%% file: main.tex
\documentclass[10pt,twocolumn,letterpaper]{article}

\usepackage[pagenumbers]{iccv} 


\definecolor{iccvblue}{rgb}{0.21,0.49,0.74}
\usepackage[pagebackref,breaklinks,colorlinks,allcolors=iccvblue]{hyperref}
\usepackage{multirow}
\usepackage{gensymb}

\def\paperID{8554} 
\def\confName{ICCV}
\def\confYear{2025}

\title{PUMPS: Skeleton-Agnostic Point-based Universal Motion Pre-Training for Synthesis in Human Motion Tasks}

\author{Clinton Ansun Mo\textsuperscript{1,2},
Kun Hu\textsuperscript{3,}\thanks{Corresponding author.},
Chengjiang Long\textsuperscript{4}, 
Dong Yuan\textsuperscript{1},
Wan-Chi Siu\textsuperscript{5},
Zhiyong Wang\textsuperscript{1} \\
\textsuperscript{1}{School of Computer Science, The University of Sydney, NSW 2006, Australia} \\
\textsuperscript{2}{The University of Tokyo, Bunkyo City, Tokyo, Japan}\\
\textsuperscript{3}{School of Science, Edith Cowan University, WA 6027, Australia}\\
\textsuperscript{4}{Meta Reality Labs, Burlingame, CA, USA} \\
\textsuperscript{5}{Hong Kong Polytechnic University, Hong Kong, China} \\
{\tt\small clinton.mo@weblab.t.u-tokyo.ac.jp, k.hu@ecu.edu.au, 
clong1@meta.com} \\
{\tt\small \{dong.yuan, zhiyong.wang\}@sydney.edu.au, enwcsiu@polyu.edu.hk}
}

\begin{document}
\maketitle
\input{sec/0_abstract}
\input{sec/1_intro}
\input{sec/2_relatedworks}
\input{sec/3_method}
\input{sec/4_experiment}
\input{sec/5_conclusion}

\section*{Acknowledgements}
This work was supported by the Australian Research Council (ARC) Linkage Project \#LP230100294, ARC Discovery Project \#DP210102674 and ECU
Science Early Career and New Staff Grant Scheme.


\input{bibliography.bbl}

\end{document}

%% file: sec/0_abstract.tex
\begin{abstract}
Motion skeletons drive 3D character animation by transforming bone hierarchies, but differences in proportions or structure make motion data hard to transfer across skeletons, posing challenges for data-driven motion synthesis.
Temporal Point Clouds (TPCs) offer an unstructured, cross-compatible motion representation. Though reversible with skeletons, TPCs mainly serve for compatibility, not for direct motion task learning.
Doing so would require data synthesis capabilities for the TPC format, which presents unexplored challenges regarding its unique temporal consistency and point identifiability.
Therefore, we propose PUMPS, the primordial autoencoder architecture for TPC data. 
PUMPS independently reduces frame-wise point clouds into sampleable feature vectors, from which a decoder extracts distinct temporal points using latent Gaussian noise vectors as sampling identifiers. 
We introduce linear assignment-based point pairing to optimise the TPC reconstruction process, and negate the use of expensive point-wise attention mechanisms in the architecture. 
Using these latent features, we pre-train a motion synthesis model capable of performing motion prediction, transition generation, and keyframe interpolation. 
For these pre-training tasks, PUMPS performs remarkably well even without native dataset supervision, matching state-of-the-art performance. When fine-tuned for motion denoising or estimation, PUMPS outperforms many respective methods without deviating from its generalist architecture. The code is available at: \url{https://github.com/MiniEval/PUMPS}.
\end{abstract}

%% file: sec/1_intro.tex
\begin{figure}
    \centering
    \includegraphics[width=\linewidth]{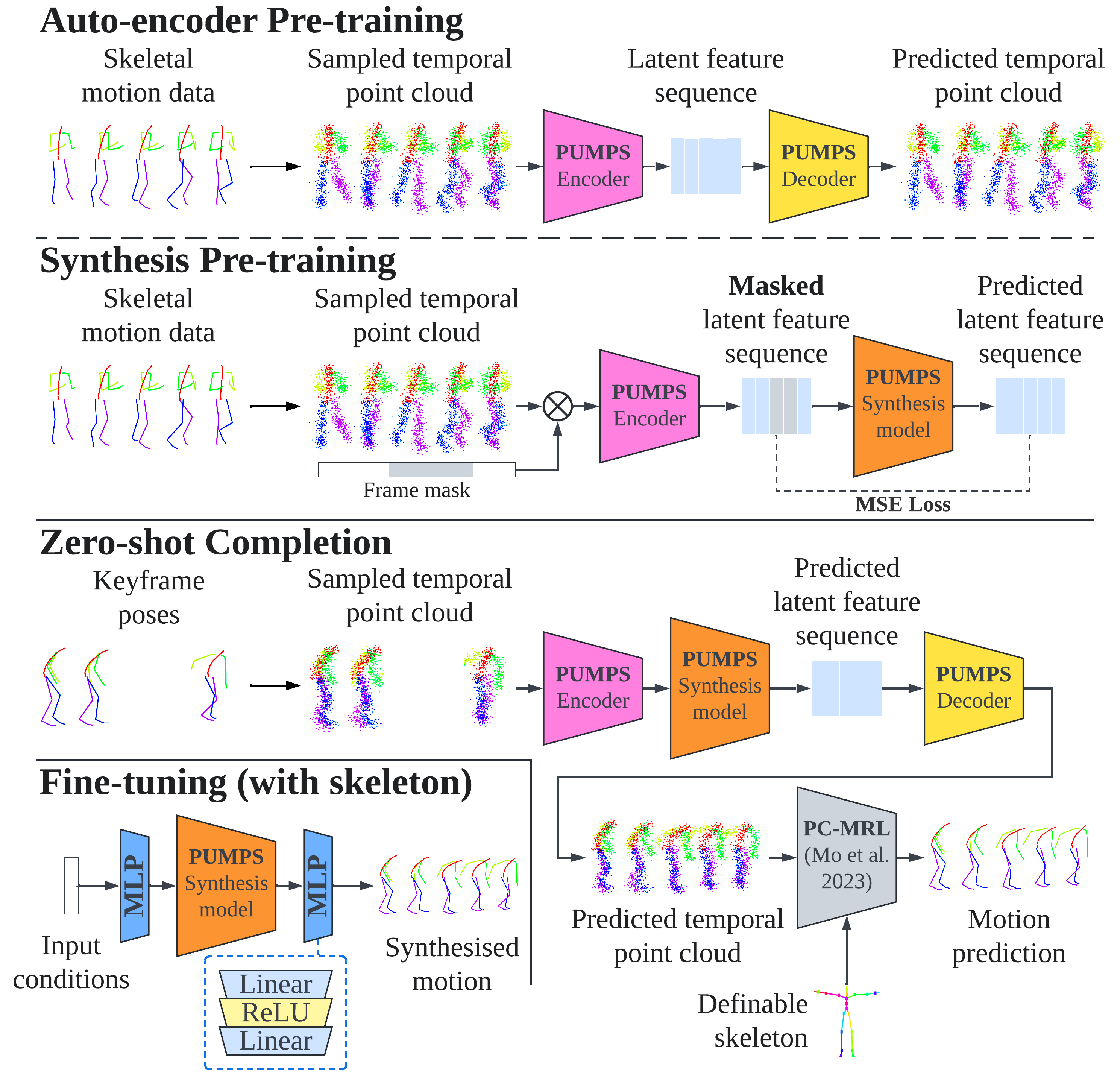}
    \caption{Overview of PUMPS pre-training, zero-shot evaluation, and fine-tuning pipelines. PUMPS consists of an auto-encoder (encoder-decoder modules) and latent synthesis component, which are pre-trained successively. 
    }
    \label{fig:intro}
\end{figure}

\section{Introduction}
\label{sec:intro}

3D motion skeletons serve as the backbone of computer animation pipelines for character motions. A skeleton is a hierarchical structure of joints in 3D space, used to deform the vertices of character meshes and other 3D features through positional and rotational transforms \citep{magnenat1988joint}. By representing joints as line segments, i.e. as \textit{bones}, skeletons provide an intuitive interface for controlling character poses. However, the diversity of skeletal configurations, with varying joint hierarchies and resting positions, has been a long-standing challenge for data-driven motion learning \citep{lee2023same}. In particular, the lack of standardization for skeletons disables machine learning methods from generalizing across different characters, ultimately limiting their applicability in animation pipelines that often target specific character setups. This is demonstrated by the joint position and rotation-based output format used conventionally in deep motion synthesis  architectures, such as for motion completion \citep{duan2022bert, mo2023continuous, oreshkin2023delta}, and conditional motion generation \citep{wang2022humanise, jin2024act}. Attempts to standardize skeleton control systems \citep{SMPL:2015, STAR:2020, AMASS:ICCV:2019} have largely traded off flexibility and expressiveness, making them less suitable for creative workflows. This limitation underscores the need for a \textit{non-intrusive} approach to motion standardisation, which does not conflict with existing character animation processes.


The concept of the \textbf{Temporal Point Cloud} (TPC) recently emerged as a geometric motion representation that obfuscates hierarchical differences of skeletons.
Compared to standard point clouds, TPCs maintain consistent point identities over a temporal dimension, allowing individual points to represent trajectories that reflect the dynamics of a motion. Crucially, trajectories allow TPCs to reconstruct skeletal motion data \textit{without supervision} \citep{mo2024pointcloud}.
While TPCs have enabled cross-skeleton motion transfer, the motion synthesis pipelines remain skeleton-specific, requiring motion synthesis to be learned independently for any new skeleton. This would be suboptimal and redundant if a unified motion synthesis solution can be achieved within the TPC medium natively, which, by definition, requires \textit{TPC synthesis}. However, TPC synthesis presents significant challenges due to the curse of dimensionality from combining spatial and temporal dimensions. 
Temporal awareness is necessary for operating with motion features,
whereas spatial awareness enables the production of evenly distributed points as expected of Gaussian-sampled TPC poses. 
Conventional architectures for capturing spatio-temporal awareness, such as attention mechanisms, are either hindered by \textit{excessive memory footprints} due to large quantities of feature vectors, or exhibit \textit{high parameter redundancy} from improper handling of point permutations.

To address these issues and enable motion feature learning without skeletal representations, we propose Point-based Universal Motion Pre-training for Synthesis (\textbf{PUMPS}) - 
an efficient two-stage pipeline for pre-training various motion tasks directly within the TPC medium. 
As illustrated in Fig. \ref{fig:intro}, PUMPS begins with an auto-encoder stage that learns a latent representation of point cloud frames, and reconstructs TPCs from these latent features. We employ conventional point attention networks \citep{zhao2021point} for the encoding stage, and introduce a novel decoder architecture tailored for TPC reconstruction.
Its key innovation is its incorporation of Gaussian noise vectors, which induce point identity-awareness into the decoding process, achieving efficient trajectory-centric TPC generation without point-wise attention. 
This practice grants our approach a significantly more simplified yet capable decoding process, with highly efficient utilisation of memory and parameters.

We leverage the resulting latent embedding space to pre-train the motion synthesis model of PUMPS, using masked data modelling tasks including \textit{keyframe interpolation}, \textit{motion transition}, and \textit{short-term motion prediction} \citep{duan2022bert, mo2023continuous}. 
Our pre-trained model achieves state-of-the-art motion completion capabilities and is capable of zero-shot motion completion for any skeleton through the TPC-to-skeleton conversion \cite{mo2024pointcloud}. Moreover, it can be effectively fine-tuned for various skeleton-specific motion tasks. We provide demonstrations of fine-tuning capabilities with \textit{2D-to-3D motion estimation} and \textit{motion denoising}.

In summary, we present the following contributions:
\begin{enumerate}
    \item We propose motion synthesis pre-training in the TPC medium, enabling universal human motion learning in a skeleton-agnostic manner.
    \item To address memory and redundancy issues with TPC reconstruction, a noise vector-based architecture is proposed to imitate point cloud sampling processes and avoid expensive spatio-temporal attention mechanisms.
    \item We demonstrate state-of-the-art motion completion capabilities of our approach through extensive experiments, as well as its adaptability in two fine-tuned tasks: \textit{motion denoising} and \textit{2D-to-3D motion estimation}.
\end{enumerate}

%% file: sec/2_relatedworks.tex
\section{Related Work}
\label{sec:related}

This section reviews existing research on machine learning approaches in the point cloud domain, along with common tasks and methods for motion synthesis in the current field. 
Since our proposed method aims to be readily usable in practice, we incorporate an appropriate variety of these synthesis tasks into the model's pre-trained capabilities.

\subsection{Permutation-Invariant Learning for Points}

Point cloud data, as an unordered set of 3D vectors, presents a unique challenge in neural network design: the requirement for permutation invariance, where models should be order-insensitive. 
Early attempts approximated point clouds as volumes \citep{wu20153d, qi2016volumetric}, leveraging order-insensitive voxel aggregations to feed into neural networks. However, volumetric representations universally suffer from resolution limitations due to their dense structure. In response, attention mechanisms on randomly-ordered sets were demonstrated on recurrent networks \citep{vinyals2016order} as a method to directly operate on set data, though consistent ordering was still preferred.
A breakthrough occurred with the development of pooling-based attention architectures, which performed true permutation-invariant aggregation using maximum or average pooling.\citep{qi2017pointnet, qi2017pointnet++, qian2022pointnext,su2025ri}.
Subsequently, Transformer-based architectures \citep{zhao2021point, wu2022ptv2, wu2024ptv3} further expanded the range of learnable point-based tasks, including point cloud completion \citep{yu2022point,wu2025dc} and point-based 3D shape synthesis \citep{wu2024ppt}.

Despite these advances, reconstructing point clouds from latent features has remained unexplored for temporally-consistent points. 
Tokenisation-based reconstruction \citep{yu2022point, pang2022masked, xu2024unipvu} lacks point identities upon which temporal points can be conditioned, whereas 2D-to-3D point folding \citep{yang2018foldingnet, quach2020folding, pang2021tearingnet} struggles to represent volumes, which are crucial for obfuscating skeletal structures. 
Our method addresses this limitation by leveraging a sequence-wise noise vector mechanism to condition and stochastically generate temporal points independently and in a volumetric manner.

\subsection{Data-Driven Motion Modelling and Synthesis}

Motion synthesis encompasses all tasks that produce motions based on a variety of conditions. Deep learning has established itself as the premier approach in this field, and primarily takes two forms: statistical motion prior-based optimisation \citep{tiwari2022posendf, zhang2021learning, peng2021amp}, and end-to-end motion synthesis learning \citep{harvey2018recurrent, petrovich2021actor, duan2022bert, tang2023predicting, oreshkin2023delta, mo2023continuous, rajendran2024review}. Motion priors model natural motion dynamics, and serve as the optimisation objective in synthesis problems with initial estimates, such as \textit{motion denoising} from noisy motions \cite{tiwari2022posendf, he2024nrdf}, and \textit{motion completion} starting from classical interpolation \citep{chai2007constraint, li2021task, tiwari2022posendf}. In contrast, end-to-end motion learning can process any synthesis condition, including those without initial estimates, such as \textit{3D pose/motion estimation} from 2D keypoints or videos \citep{martinez2017simple, li2022mhformer, zheng2021poseformer, zhao2023poseformerv2}, \textit{sign language translation} \citep{kan2022sign,xue2023alleviating,yu2024signavatars}, and \textit{text-to-motion generation} \citep{kim2023flame, zhang2024motiondiffuse, wang2023multi}. Unlike motion priors, end-to-end learning generally requires training data that specifically pertains to the synthesis task. As end-to-end solutions become increasingly prominent in motion synthesis, the necessity of compatible task-specific motion datasets stands out as a major obstacle towards their practicalities outside of lab environments.


%% file: sec/3_method.tex
\section{PUMPS Architecture}

PUMPS consists of three components: 1) \textit{Point cloud frame encoder} $\Phi^\text{enc}$ to transform TPCs into a rich latent space, 2) \textit{TPC reconstruction decoder} $\Phi^\text{dec}$ to reconstruct temporal-aware point clouds, and 3) \textit{Latent motion synthesiser} $\Phi^\text{LMS}$ to learn motion synthesis through masked self-supervision.

\begin{figure}
    \centering
    \includegraphics[width=\linewidth]{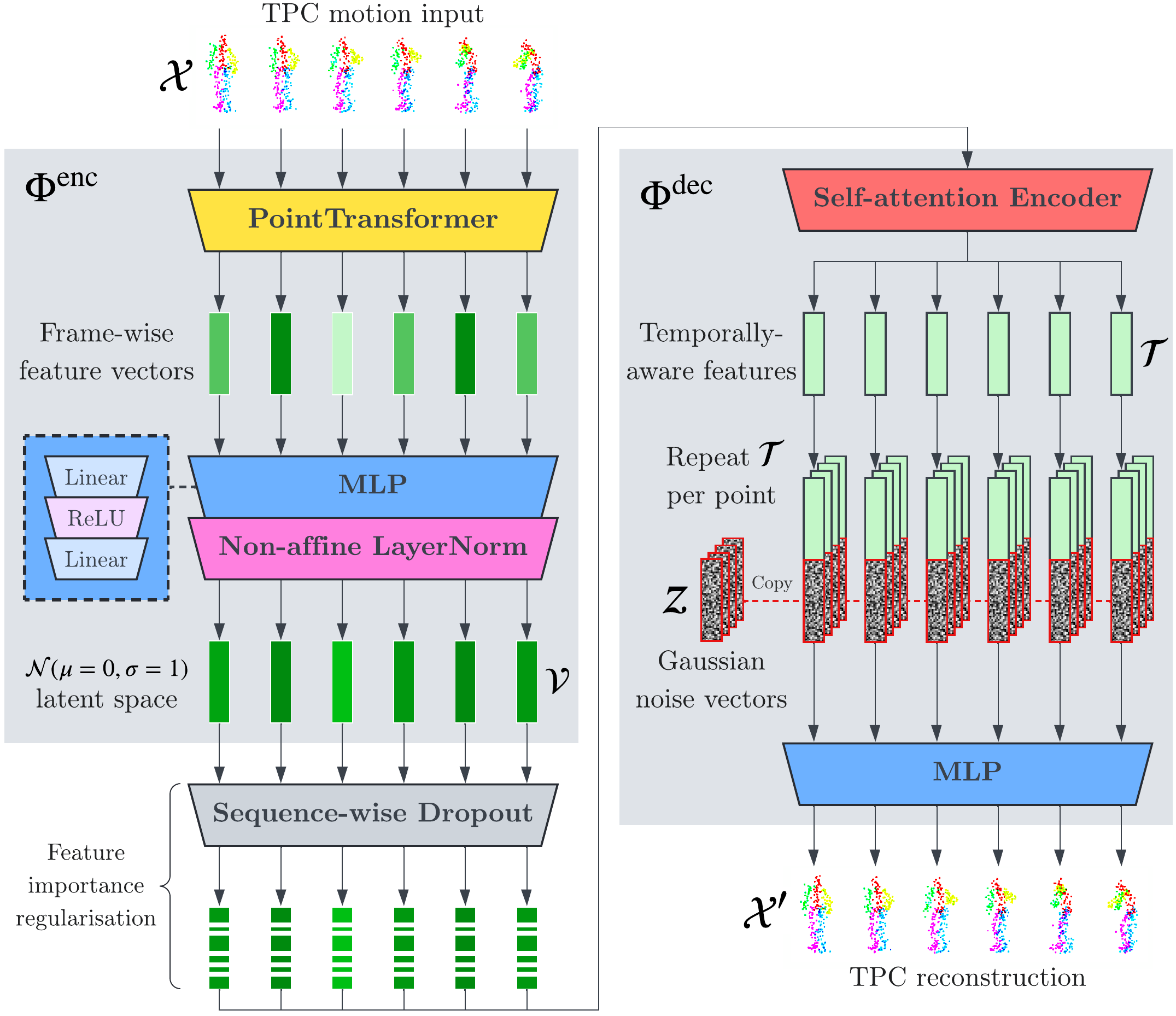}
    \caption{Training strategy of the PUMPS auto-encoder components $\Phi^\text{enc}$ and $\Phi^\text{dec}$. The primary objectives of the auto-encoder are to learn a TPC reconstruction process for $\mathcal{P}'$, and produce a regularised latent space $\mathcal{V}$ of point cloud frame features.} 
    \label{fig:enclmm}
\end{figure}

\subsection{Point Cloud Frame Encoder}

The frame encoder $\Phi^\text{enc}$ is a crucial component of PUMPS, designed to extract rich latent features of pose patterns from individual point cloud frames sampled from skeletal poses. 
By using point clouds as a universally compatible representation, the encoder can be trained on various motion datasets, regardless of their native skeleton structure. 
The resulting skeleton-agnostic latent space serves as a foundation for learning a wide range of motion synthesis tasks. 

To begin, we generate point cloud data from the proximate volumes of posed skeletons using the TPC sampling setup in \citep{mo2024pointcloud}. 
Formally, given a pose $x$, we can sample its point cloud representation as $\mathcal{P}_x = \{p_0, p_1, ..., p_{|\mathcal{P}| - 1}\}$, where $|\mathcal{P}|$ indicates the number of points in the point cloud.
Each point $p \in \mathcal{P}$ consists of a 3D position and a one-hot body group based on the bone with which it is associated, assigned to one of the following classes: \{\textit{body, left arm, right arm, left leg, right leg}\}. 
The encoder converts $\mathcal{P}_x$ into a $|v|$-element vector $v_x \in \mathbb{R}^{|v|}$. For point aggregation, our setup trains a PointTransformer backbone \citep{zhao2021point} from scratch.

$\Phi^\text{enc}$ is trained using an auto-encoder objective: reconstructing the input point cloud from its latent representation (discussed in Sec. \ref{sec:reconstruct}).
To make reconstructions of latent vector $v$ learnable with scalar error-based objectives (in Sec. \ref{sec:lms}),
the latent space must 1) maintain a consistent distribution between features and 2) ensure near-equal importance for each feature in the decoding process.
In detail, we employ \textit{non-affine layer normalisation} after $\Phi^\text{enc}$ to restrain features within a normal $\mathcal{N}(0, 1)$ distribution, and a \textit{sequence-wise dropout} strategy in $\Phi^\text{dec}$ to regularise individual feature importances. 
Sequence-wise dropout omits the same feature indices throughout a given sequence, ensuring that remaining features retain continuity, as opposed to frame-independent omission with standard dropout.

\subsection{Point Cloud Reconstruction Decoder}
\label{sec:reconstruct}

TPCs present several significant challenges for the reconstruction process. 
A TPC decoder should account for point identities to maintain temporal consistency when tracking points across frames. 
Intuitively, spatio-temporal self-attention could enable conditioning on both point identity and position; however, this approach is prohibitively expensive due to the quadratic complexity of self-attention. Given an encoded latent vector sequence $\mathcal{V} = \{v_0, v_1, ..., v_{|\mathcal{V}|-1}\}$, spatio-temporal attention would result in a complexity of $O((|\mathcal{P}||\mathcal{V}|)^2)$, where $|\mathcal{V}|$ is the frame length of the motion. 

We propose the construction of a latent space sampling system that can extract and reconstruct \textit{independent points} from latent features, eliminating the need for costly point-wise attention. 
The architecture of $\Phi^\text{dec}$, illustrated in Fig. \ref{fig:enclmm}, is designed to reconstruct $|\mathcal{P}|$ point sequences from the latent vector sequence $\mathcal{V}$. Since each $v \in \mathcal{V}$ is derived independently, without temporal awareness, $\Phi^\text{dec}$ uses rotary self-attention (RoPE) \citep{su2024roformer} to produce a temporally-aware transformation $\mathcal{T} = \{t_0, t_1, ..., t_{|\mathcal{V}|-1}\}$. 

\begin{figure}
    \centering
    \includegraphics[width=\linewidth]{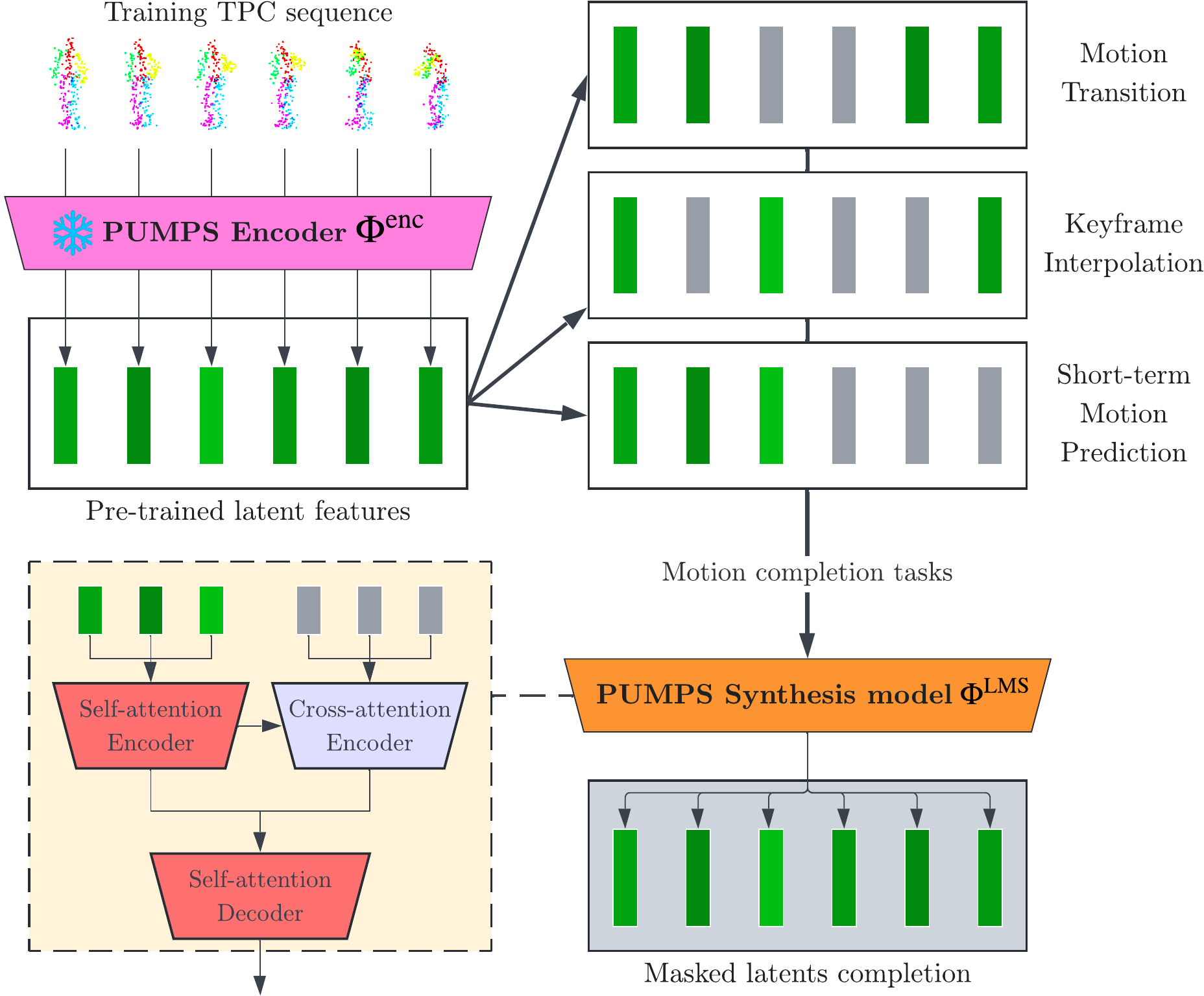}
    \caption{Motion completion pre-training for the latent motion synthesiser $\Phi^\text{LMS}$, using a frozen pre-trained checkpoint for $\Phi^\text{enc}$. Gray latent feature vectors are masked out from the $\Phi^\text{LMS}$ input.}
    \label{fig:lms}
\end{figure}

To produce distinct temporal points for reconstructing the TPC, 
we introduce stochasticity via a $|z|$-element Gaussian noise vector $z \in \mathbb{R}^{|z|} \sim \mathcal{N}(0, 1)$ for each point, which is appended to each latent vector in $\mathcal{T}$. This results in a set of $|\mathcal{P}| \times |\mathcal{V}|$ vectors: $\{t_0 \oplus z_0, t_1 \oplus z_0, t_2 \oplus z_0, ..., t_0 \oplus z_1, t_1 \oplus z_1, t_2 \oplus z_1, ...\}$, where $t_\alpha \oplus z_\beta$ represents the $\alpha$-th frame of the $\beta$-th point. Each point maintains a consistent identity across different frames by having the same $z_\beta$.
In addition, we condition the body group of each point, similarly to the encoding process of $\Phi^\text{enc}$. 
Finally, each vector of this vector set is independently decoded into a 3D position sequence using multi-layer perceptron (MLP) layers, producing a reconstructed TPC. 
By deferring attention mechanisms before expanding to the point dimension, $\Phi^\text{dec}$ achieves a complexity of $O(|\mathcal{V}|^2)$ for the RoPE attention step, followed by a set of $O(|\mathcal{P}||\mathcal{V}|)$ independent attention-free operations.

The absence of point attention causes a lack of inter-point awareness in $\Phi^\text{dec}$. Consequently, common Chamfer distance-based objectives \citep{wu2021densityaware, yu2022point, mo2024pointcloud} often cause predicted points to collapse toward the centroid of body group clusters. 
This is because Chamfer distance addresses the presence, but not the \textit{expected distribution}, of points within the reconstructed cloud. 
We propose the use of \textbf{linear assignment}. 
Given the expected and predicted TPCs $\mathcal{P}$ and $\mathcal{P}'$, assignments methods like the Hungarian method \citep{edmonds1972theoretical} can deduce the optimal set of unique point pairs with the minimum total distance measured in Mean Squared Error (MSE). We denote the pairing set as $H(\mathcal{P}, \mathcal{P}') = \{(p_0, p'_{h_0}), (p_1, p'_{h_1}), ...\}$, where $\{h_0, h_1, ..., h_{|\mathcal{P}| - 1}\} = \{0, 1, ..., |\mathcal{P}| - 1\}$, 
from which $\Phi^\text{enc}$ and $\Phi^\text{dec}$ can optimise the reconstruction objective $\mathcal{L}_\text{rec}$: 
\[\mathcal{L}_\text{rec} = \sum_{0 \leq k < |\mathcal{P}|}||p_k - p'_{h_k}||_2^2.\]
\subsection{Latent Motion Synthesiser}
\label{sec:lms}

We train our motion synthesis (pre-training) tasks on an existing Transformer-based architecture, which was proposed for motion interpolation learning \citep{mo2023continuous}, denoted as $\Phi^\text{LMS}$. 
Latent encodings of the pose data are created by a frozen $\Phi^\text{enc}$ checkpoint. 
The tasks involve \textit{keyframe interpolation}, \textit{motion transition}, and \textit{short-term motion prediction}, which can all be defined as masked data modelling problems, as illustrated in Fig. \ref{fig:lms}. 
Formally, our masked data modelling involves predicting an expected $|\mathcal{V}|$-frame latent sequence $\mathcal{V} = \{v_0, v_1, ..., v_{|\mathcal{V}|-1}\}$ from a subset of given data $\mathcal{V}' = \{v_{k_0}, v_{k_1}, ...\}$, determined by a keyframe set $K = \{k_0, k_1, ...\}, K \subset \{0, 1, 2, ..., |\mathcal{V}|-1\}$. 

\begin{itemize}
    \item For \textbf{keyframe interpolation} learning, all $|K|$ keyframes are randomly selected, except the first and last frames which are always selected, i.e. $K = \{0, |\mathcal{V}|-1\} \cup \{\mathbb{N}^{|K| - 2} \sim \mathcal{U}_{[1, |\mathcal{V}|-2] \cap \mathbb{N}}\}$.
    \item For \textbf{motion transition} learning, keyframes produce a masked interval between two known motion sequences, i.e. $K = \{k \in \mathbb{N} \mid 0 \leq k < |\mathcal{V}|\} \setminus \{k \in \mathbb{N} \mid a \leq k \leq b, a > 0, b < |\mathcal{V}|-1\}$.
    \item For \textbf{short-term motion prediction} learning, a continuous segment is removed from the end of the sequence, i.e. $K = \{k \in \mathbb{N} \mid 0 \leq k < b, b < |\mathcal{V}|-1\}$.
\end{itemize}

To optimise $\Phi^\text{LMS}$, we minimise the MSE between $\mathcal{V}$, and predicted latent vectors $\mathcal{W} = \{w_0, w_1, ..., w_{|\mathcal{V}|-1}\}$, i.e. $\mathcal{L}_\text{LMS} = \sum_{0 \leq f < |\mathcal{V}|} ||v_f - w_f||_2^2$. Normalisation steps in $\Phi^\text{enc}$ are critical to ensuring consistent variances between features, validating the use of MSE as an objective.

%% file: sec/4_experiment.tex
\begin{table*}
    \centering
    \setlength\tabcolsep{3.5px}
    \resizebox{\linewidth}{!}{
        \begin{tabular}{r c c c c c c c c c r c c c c c c c c c}
            \textit{Keyframe} & \multicolumn{3}{c}{\textbf{L2P}$\downarrow$} & \multicolumn{3}{c}{\textbf{L2Q}$\downarrow$} & \multicolumn{3}{c}{\textbf{NPSS}$\downarrow$} & \textit{Motion} & \multicolumn{3}{c}{\textbf{L2P}$\downarrow$} & \multicolumn{3}{c}{\textbf{L2Q}$\downarrow$} & \multicolumn{3}{c}{\textbf{NPSS}$\downarrow$} \\
            \textit{interpolation} & 5 & 15 & 30 & 5 & 15 & 30 & 5 & 15 & 30 & \textit{transition} & 15 & 30 & 60 & 15 & 30 & 60 & 15 & 30 & 60 \\
            \cmidrule(lr){1-1} \cmidrule(lr){2-4} \cmidrule(lr){5-7} \cmidrule(lr){8-10} \cmidrule(lr){11-11}  \cmidrule(lr){12-14} \cmidrule(lr){15-17} \cmidrule(lr){18-20}
            LERP & 0.170 & 0.588 & 1.166 & 0.285 & 0.765 & 1.173 & 0.354 & 1.333 & 2.420 & LERP & 0.248 & 0.743 & 1.889 & 0.294 & 0.673 & 1.266 & 0.248 & 1.098 & 2.916 \\
            SLERP & 0.170 & 0.591 & 1.172 & 0.286 & 0.766 & 1.181 & 0.357 & 1.369 & 2.488 & SLERP & 0.249 & 0.745 & 1.896 & 0.292 & 0.679 & 1.266 & 0.243 & 1.100 & 2.953 \\
            $\text{TG}_\text{complete}$ & 0.169 & 0.542 & 1.138 & 0.283 & 0.728 & 1.182 & 0.351 & 1.202 & 2.200 & $\text{TG}_\text{complete}$ & 0.222 & 0.676 & 1.913 & 0.279 & 0.663 & 1.206 & \textbf{0.216} & 1.030 & 2.579 \\
            BERT & 0.203 & 0.493 & 0.998 & 0.325 & 0.736 & 1.170 & 0.541 & 1.299 & 2.213 & BERT & 0.201 & 0.517 & 1.331 & 0.296 & 0.656 & 1.165 & 0.226 & 1.049 & \textbf{2.127} \\
            Pose-NDF & 0.206 & 0.609 & 1.180 & 0.322 & 0.792 & 1.196 & 0.369 & 1.350 & 2.441 & Pose-NDF & 0.304 & 0.789 & 1.925 & 0.360 & 0.747 & 1.328 & 0.255 & 1.127 & 2.954 \\
            MAE & 0.182 & 0.502 & 0.978 & 0.302 & 0.739 & 1.128 & 0.471 & 1.316 & 2.192 & MAE & 0.188 & 0.456 & 1.234 & 0.283 & 0.644 & \textbf{1.159} & 0.221 & 1.004 & 2.169 \\
            $\Delta$ & 0.154 & 0.476 & 0.941 & 0.263 & \textbf{0.674} & 1.058 & 0.322 & 1.170 & 2.073 & $\Delta$ & 0.239 & 0.658 & 1.707 & 0.274 & 0.631 & 1.168 & 0.218 & 1.018 & 2.691 \\
            CITL & 0.186 & 0.466 & \textbf{0.908} & 0.307 & 0.688 & \textbf{1.056} & 0.536 & 1.271 & 2.023 & CITL & 0.255 & 0.632 & 1.656 & 0.334 & 0.718 & 1.230 & 0.269 & 1.117 & 2.300 \\
            \cmidrule(lr){1-1} \cmidrule(lr){11-11}
            \textbf{PUMPS} & \textbf{0.137} & \textbf{0.458} & 0.948 & \textbf{0.260} & 0.693 & 1.095 & \textbf{0.249} & \textbf{1.080} & \textbf{1.973} & \textbf{PUMPS} & \textbf{0.168} & \textbf{0.443} & \textbf{0.900} & \textbf{0.271} & \textbf{0.609} & 1.223 & 0.241 & \textbf{0.946} & 2.766 \\
            \cmidrule(lr){1-10} \cmidrule(lr){11-20} 
        \end{tabular}
    }
    \caption{Keyframe interpolation (left) and motion transition (right) $\ell_2$ position, $\ell_2$ quaternion, and NPSS performance. Keyframe interpolation is tested on 121-frame motion sequences, given regular 5, 15, and 30-frame intervals between keyframes. Motion transition is tested on 15, 30, and 60-frame intervals between 15-frame motion windows. The top performer for each scenario is highlighted in \textbf{bold}.}
    \label{tab:interp}
\end{table*}

\section{Masked Motion Synthesis Pre-Training}
\label{sec:pretraining}

\subsection{Motion Datasets}

We evaluate the effectiveness of our proposed method on the pre-training motion completion tasks, comparing it to existing methods in the field. 
The data we use for training and evaluation are sourced from the following datasets:

\begin{itemize}
    \item \textbf{AMASS} \citep{AMASS:ICCV:2019} is a large collection of motion datasets retargeted to various SMPL skeletons \citep{SMPL:2015} via manual vertex mesh-based weighting and optimisation.
    \item \textbf{LaFAN1} \citep{harvey2020robust} consists of production quality, long-form human motions for dynamic action classes commonly used in video games, such as \textit{firearm aiming}.
    \item \textbf{Human3.6M} \citep{IonescuSminchisescu11, h36m_pami} contains a large set of natural human motions, captured on a compact stage. This dataset generally features slower, more static movements.
    \item \textbf{100STYLE} \citep{mason2022local} contains 100 distinct motion styles focused on locomotion, such as walking and running.
\end{itemize}

We designate all datasets for pre-training PUMPS, excluding testing splits from LaFAN1 and 100STYLE.

\subsection{Experimental Settings}

PUMPS operates on TPCs consisting of 256 points, sampled from skeletons using a distribution of $\sigma \sim \mathcal{U}_{[0.025, 0.075]}$ for point offsets from their associated bones \citep{mo2024pointcloud}. The encoder $\Phi^\text{enc}$ initially learns feature vectors of size 32 per point. 
$\Phi^\text{enc}$ employs 4 layers of farthest-first pooling \citep{yu2022point}, with each layer doubling the feature size while reducing the ratio of point vectors by $\frac{1}{4}$, resulting in a singular feature vector of size 512. 
For the point cloud decoder $\Phi^\text{dec}$, we employ 4 layers of RoPE self-attention on the latent sequence, and 3 linear MLP layers to reduce point feature sizes to 512, 32, and 3 respectively. A dropout ratio of 0.1 is used for the sequence-wise dropout mechanism.
$\Phi^\text{LMS}$ uses 8 attention heads and layers for each attention stage. 
For training, we employ an AdamW optimiser \citep{loshchilov2019decoupled}, using a learning rate of $10^{-4}$ with exponential decay \citep{vaswani2017attention}. 

Each pre-training task is configured as follows, on 128-frame motion sequences:
\begin{itemize}
    \item \textbf{Keyframe interpolation}: $5 \leq |K| \leq 32$ keyframes are given as input, and missing frames are predicted.
    \item \textbf{Motion transition}: A contiguous masked section of $10 \leq 128 - |K| \leq 32$ frames is to be predicted.
    \item \textbf{Motion prediction}: The last $10 \leq 128 - |K| \leq 32$ frames are to be predicted.
\end{itemize}

We compare our keyframe interpolation and motion transition results with several state-of-the-art machine learning methods, as well as standard linear and spherical interpolation (LERP \& SLERP) techniques \citep{shoemake1985slerp}. 
The learned methods include the recurrent network-based $\text{TG}_\text{complete}$ architecture \citep{harvey2020robust}, a prior-based interpolation approach PoseNDF \citep{tiwari2022posendf}, and several attention-based approaches, namely a BERT adaptation \citep{duan2022bert}, $\Delta$-interpolator \citep{oreshkin2023delta}, masked auto-encoders (MAE) \citep{he2022masked}, and CITL \citep{mo2023continuous}. 
We measure accuracy based on $\ell_2$ distances between global bone positions and quaternions (L2P 
\& L2Q), as well as the signal similarity metric, NPSS \citep{gopalakrishnan2019neural}. 
To retrieve skeletal motion data from our TPCs, we adapt a RoPE variant of the unsupervised point cloud-to-skeletal motion model, PC-MRL, from \citep{mo2024pointcloud}. 
To further address imperfections in motion data retrieval, we correct our model by applying offsets measured between predicted and known keyframes.

\subsection{Motion Synthesis Results}

\begin{figure*}
    \centering
    \includegraphics[width=\linewidth]{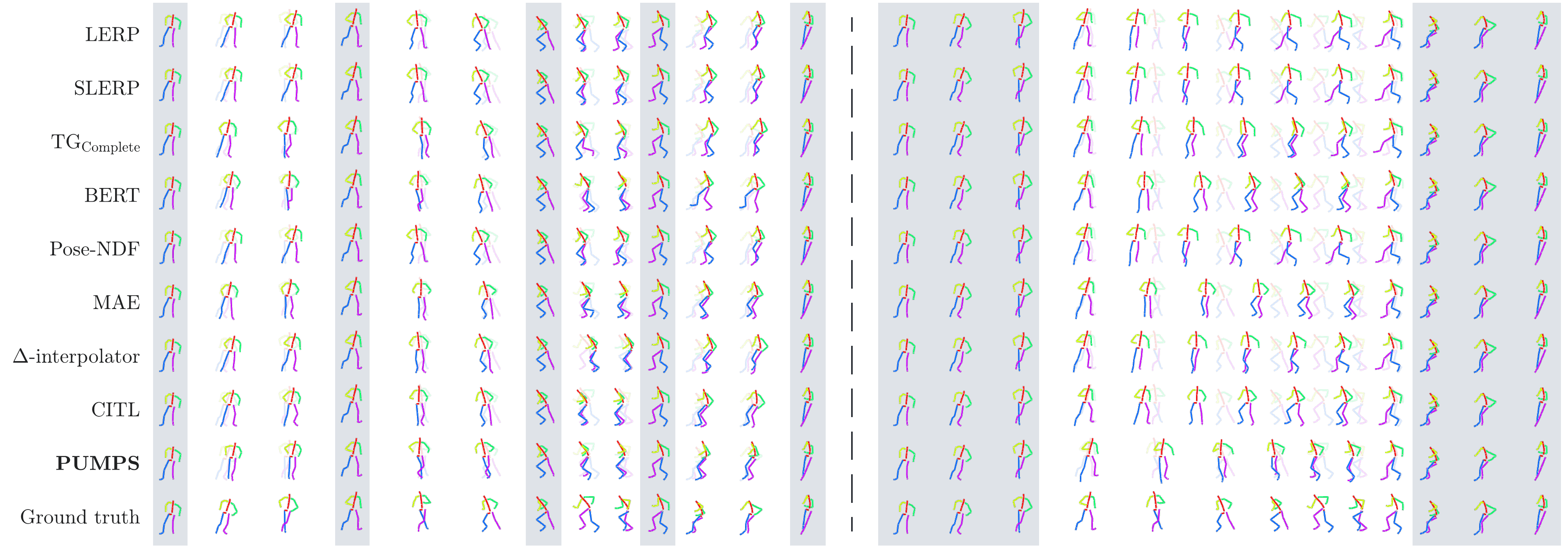}
    \caption{Keyframe interpolation (left) and motion transition (right) examples on a 61-frame sequence. We leave 15 frames between keyframes for interpolation, and 31 frames between 15-frame windows for transition. For visibility purposes, only every 5th frame is shown. Grey regions indicate frames provided by the input. The ground truth is overlaid in each row for direct comparison.}
    \label{fig:interp}
\end{figure*}

As shown in Table \ref{tab:interp}, 
PUMPS demonstrates surprisingly superior interpolation and transition performance to existing natively supervised methods, given that PUMPS relies on non-native motion data in training.
The main strength of PUMPS lies in its L2P performance, which is to be expected as point clouds directly represent positional data. 
Positional accuracy is essential for strong synthesis performance, as joint positions influence visual similarity more directly than rotations and frequencies. 
Nonetheless, the rotational representation capabilities of TPCs, as first shown in \citep{mo2024pointcloud}, remain effective with PUMPS, given its near-equal L2Q performance with existing methods. PUMPS strongly compensates its lack of native supervision with its ability to simultaneously learn from multiple large motion datasets, utilising homogenised TPC-based motion representations.

The running motion example in Fig.~\ref{fig:interp} highlights a key advantage of interpolating within our point cloud latent space compared to direct skeletal data predictions. It illustrates a sequence with motion patterns that are unlikely to be present in the training portion of the dataset. In such cases, masked motion synthesis models often default to a LERP-like strategy, which may fail to respect the expected inertial properties of movements. 
The behaviour primarily arises from LERP's status as a relatively straightforward solution for interpolating skeletal data \citep{mo2023continuous, oreshkin2023delta}, which most machine learned approaches tend to converge as default behaviour. 
In contrast, LERP does not exhibit any notable efficacy in the PUMPS latent space, as the relation between our latent features and its resulting point clouds is both non-linear and stochastic, structured by $\Phi^\text{dec}$. 
Furthermore, since PUMPS is able to learn from a significantly larger dataset than existing methods, its solution is much more generalised and exhibits inertial behaviour more strongly than other methods. The combination of these factors makes PUMPS a new state-of-the-art solution for masked motion synthesis.

\subsection{Ablation Study on TPC Reconstruction}

\begin{table}
    \centering\small
    \begin{tabular}{l c}
        \textit{Component} & $\mathcal{L}_\text{rec} \downarrow$ \\
        \hline
        Element-wise dropout \textit{(No sequence-wise dropout)} & 0.0204 \\
        Sinusoidal positions \textit{(No rotary self-attention)} & 0.0052 \\
        Chamfer distance objective \textit{(No linear assignment)} & 0.0085 \\
        Point sampling: PointIVAE \citep{chen2022pointivae} & 0.0260 \\
        Point sampling: FoldingNet \citep{yang2018foldingnet} & 0.0040 \\
        \hline
        \textbf{Our method} & \textbf{0.0029} \\
        \hline
    \end{tabular}
    \caption{Component ablation results of the PUMPS auto-encoder architecture, measured in linearly assigned point distances ($\mathcal{L}_\text{rec}$).}
    \label{tab:ablation}
\end{table}

The core technical contributions of our work are centred around TPC reconstruction. As such, in Table \ref{tab:ablation}, we demonstrate the performance contributions of the various components in the PUMPS auto-encoder. Of the basic component studies, the only alternative configuration that converged to some semblance of motion was sinusoidal positional embeddings, used in the original Transformer model \cite{vaswani2017attention}. Conventional element-wise dropout induced erratic motion dynamics that prevented convergence to a consistent motion, while Chamfer distances collapsed into distinct point clusters rather than evenly distributed point volumes.

\begin{figure}
    \centering
    \includegraphics[width=\linewidth]{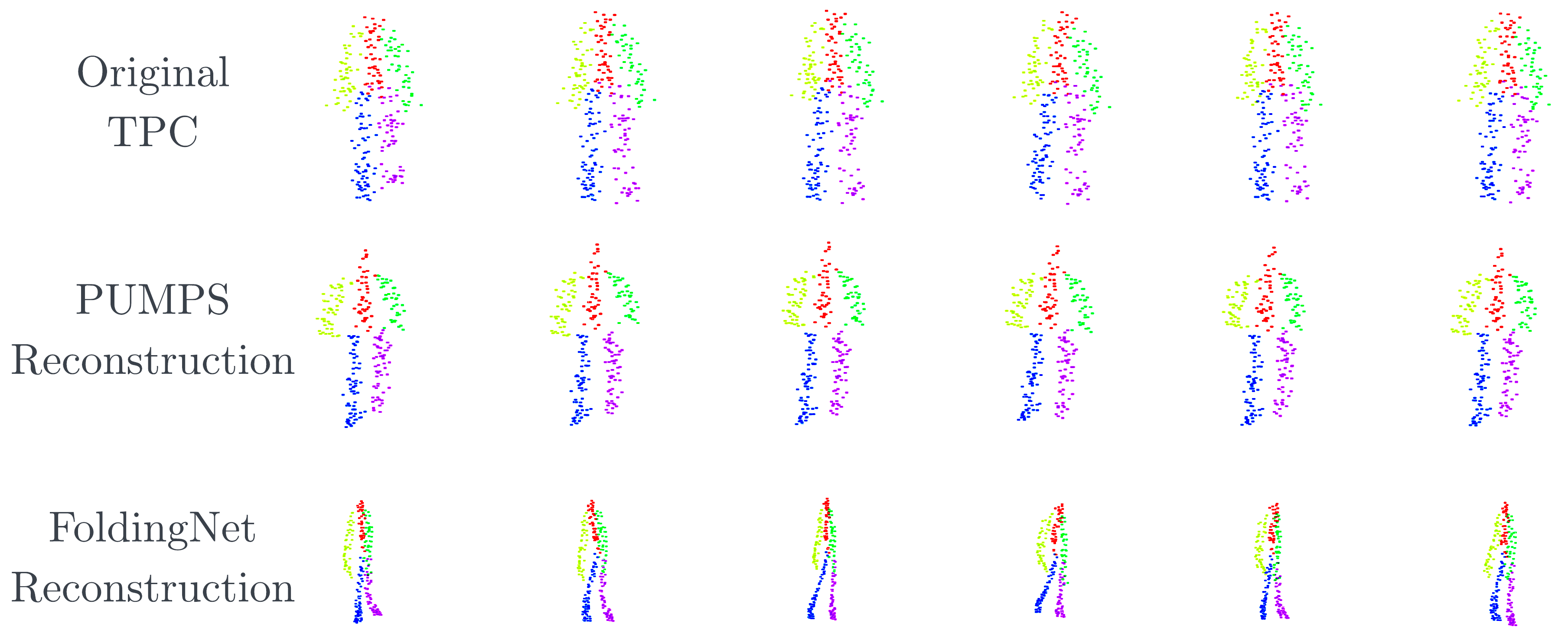}
    \caption{TPC reconstructions comparisons between our method and FoldingNet \citep{yang2018foldingnet}, from the same TPC motion representation.}
    \label{fig:foldingnet}
\end{figure}

We further showcase two existing point cloud reconstruction strategies that exhibit point identity conditioning mechanisms for TPCs: PointIVAE \cite{chen2022pointivae} with VAE-based sampling, and FoldingNet \cite{quach2020folding} with 2D-to-3D point map folding. The implementation of these components are centred around producing the TPC from $\mathcal{T}$. The volatility of latent vectors in VAE decoders under conventional Kullback-Leibler divergence regimes \citep{kingma2013auto} appears to make point cloud modelling impossible to converge with a temporal axis. On the other hand, FoldingNet \cite{yang2018foldingnet} shares a key similarity with our method in modelling \textit{point identities}, using uniform 2D coordinates rather than noise vectors, and demonstrates a reasonable level of TPC modelling capability. However, two identifier variables alone cannot reasonably produce a volumetric point cloud representation, exemplified by FoldingNet's inability to model the distributive features of TPCs, as illustrated in Fig. \ref{fig:foldingnet}.

\section{Fine-Tuning for Motion-based Tasks}
\label{sec:finetuning}

For fine-tuning demonstrations, we train skeleton-specific models for 3D motion estimation and denoising tasks. For both tasks, we include an input and output MLP layer in $\Phi^\text{LMS}$ to project the skeleton's pose space into our latent space, and vice versa, as shown in Fig. \ref{fig:intro}.

\subsection{2D-to-3D Motion Estimation}
\label{sec:estimation}

3D motion estimation from video sequences often begins with 2D keypoint estimation to  track body parts. Given 2D perspective projections of 3D skeletal motions as input, motion estimation models aim to reconstruct the original motion by predicting skeletal bone rotations. 
Using joint-wise $\ell_2$ position and quaternion objectives, we perform end-to-end fine-tuning of the pre-trained $\Phi^\text{LMS}$ model on supervised motion estimation to demonstrate the adaptability of our method. Since 2D skeletons cannot be represented by 3D point clouds, this fine-tuned model directly learns latent projections from 2D to 3D skeletal data.

We evaluate two variants of our $\Phi^\text{LMS}$ model - one fine-tuned and one trained from scratch - against multiple state-of-the-art 2D-to-3D motion estimation methods, including the spatio-temporal transformer-based models MixSTE \citep{zhang2022mixste}, PoseFormer v1/v2 \citep{zheng2021poseformer, zhao2023poseformerv2}, and MHFormer \citep{li2022mhformer}, as well as a motion pre-training approach, MotionBERT \citep{zhu2023motionbert}. 
To obtain 2D keypoints, we simply evaluate perspective projection maps of 3D joint positions from a random camera position, with a rotation that sets the camera's principal point to the skeleton's root joint position.

Motion estimation accuracy is commonly measured with mean per-joint positional error (MPJPE) and velocity error (MPJVE) \citep{IonescuSminchisescu11, zhao2023poseformerv2}, metrics that we adopt for our benchmark. 
As shown in Table \ref{tab:estimation}, our fine-tuned model produces competitive results against state-of-the-art methods in terms of both MPJPE and MPJVE, and notably outperforms its own architecture if trained from scratch. 
The respective ${\sim}40\%$ and ${\sim}30\%$ reduction in each metric strongly demonstrate the importance of the pre-training process for effective motion estimation learning. This improvement enables our method to consistently outperform all existing methods except MHFormer, which demonstrates strong generalisability through its sophisticated dropout setup. It should be noted that most existing approaches utilise spatial attention mechanisms, which is absent from our $\Phi^\text{LMS}$ model. In future work, we may consider more sophisticated fine-tuning strategies involving such mechanisms.

\begin{table}
    \setlength\tabcolsep{3px}
    \begin{subtable}[h]{\linewidth}
        \resizebox{\linewidth}{!}{
            \begin{tabular}{l c c c c c c c c}
                \multirow{2}{*}{\textbf{MPJPE}$\downarrow$} & \multicolumn{8}{c}{\textit{Motion category}} \\
                \cmidrule(lr){2-9}
                 & \textit{Aim} & \textit{Dance} & \textit{Fall} & \textit{Fight} & \textit{Obst.} & \textit{Prone} & \textit{Push} & \textit{Walk} \\
                \cmidrule(lr){1-1} \cmidrule(lr){2-9}
                MixSTE & 61.05 & 58.26 & 80.95 & 60.58 & 54.47 & 95.13 & 58.20 & 47.68 \\
                PoseFormer v1 & 71.21 & 82.34 & 92.72 & 77.35 & 72.65 & 102.71 & 71.51 & 71.90 \\
                MotionBERT & 57.80 & 59.05 & 73.86 & 59.83 & 52.27 & 76.18 & 51.74 & 48.99 \\
                MHFormer & 40.38 & 45.43 & 55.59 & 43.49 & 38.36 & 62.87 & 37.47 & 37.67 \\
                PoseFormer v2 & 59.19 & 59.80 & 87.72 & 60.01 & 48.82 & 93.57 & 49.17 & 45.32 \\
                \cmidrule(lr){1-1}
                \textbf{PUMPS} (scratch) & 73.56 & 84.98 & 99.46 & 84.39 & 67.02 & 119.80 & 65.94 & 57.79 \\
                \textbf{PUMPS} (finetune) & 44.15 & 48.63 & 57.30 & 45.12 & 37.97 & 73.26 & 38.53 & 38.65 \\
                \bottomrule
            \end{tabular}
        }
    \end{subtable}
    \begin{subtable}[h]{\linewidth}
        \centering
        \resizebox{\linewidth}{!}{
            \begin{tabular}{l c c c c c c c c}
                \toprule
                \multirow{2}{*}{\textbf{MPJVE}$\downarrow$} & \multicolumn{8}{c}{\textit{Motion category}} \\
                \cmidrule(lr){2-9}
                 & \textit{Aim} & \textit{Dance} & \textit{Fall} & \textit{Fight} & \textit{Obst.} & \textit{Prone} & \textit{Push} & \textit{Walk} \\
                \cmidrule(lr){1-1} \cmidrule(lr){2-9}
                MixSTE & 23.44 & 39.38 & 22.59 & 27.90 & 22.83 & 26.41 & 26.44 & 23.01 \\
                PoseFormer v1 & 14.92 & 36.61 & 22.16 & 27.10 & 18.16 & 19.71 & 21.81 & 18.70 \\
                MotionBERT & 18.30 & 32.99 & 25.44 & 28.53 & 18.50 & 21.28 & 23.45 & 19.72 \\
                MHFormer & 9.66 & 22.55 & 15.99 & 18.07 & 10.65 & 13.39 & 13.49 & 11.01 \\
                PoseFormer v2 & 16.64 & 30.13 & 23.52 & 26.68 & 14.66 & 19.86 & 19.55 & 14.74 \\
                \cmidrule(lr){1-1}
                \textbf{PUMPS} (scratch) & 18.69 & 42.33 & 31.60 & 35.15 & 19.84 & 27.21 & 27.19 & 19.53 \\
                \textbf{PUMPS} (finetune) & 12.65 & 27.14 & 18.37 & 19.92 & 12.91 & 16.69 & 17.96 & 15.36 \\
                \bottomrule
            \end{tabular}
        }
    \end{subtable}
    \caption{2D-to-3D motion estimation method comparisons on 128-frame sequences, using MPJPE (top) and MPJVE (bottom).}
    \label{tab:estimation}
\end{table}
\begin{figure}
    \centering
    \includegraphics[width=\linewidth]{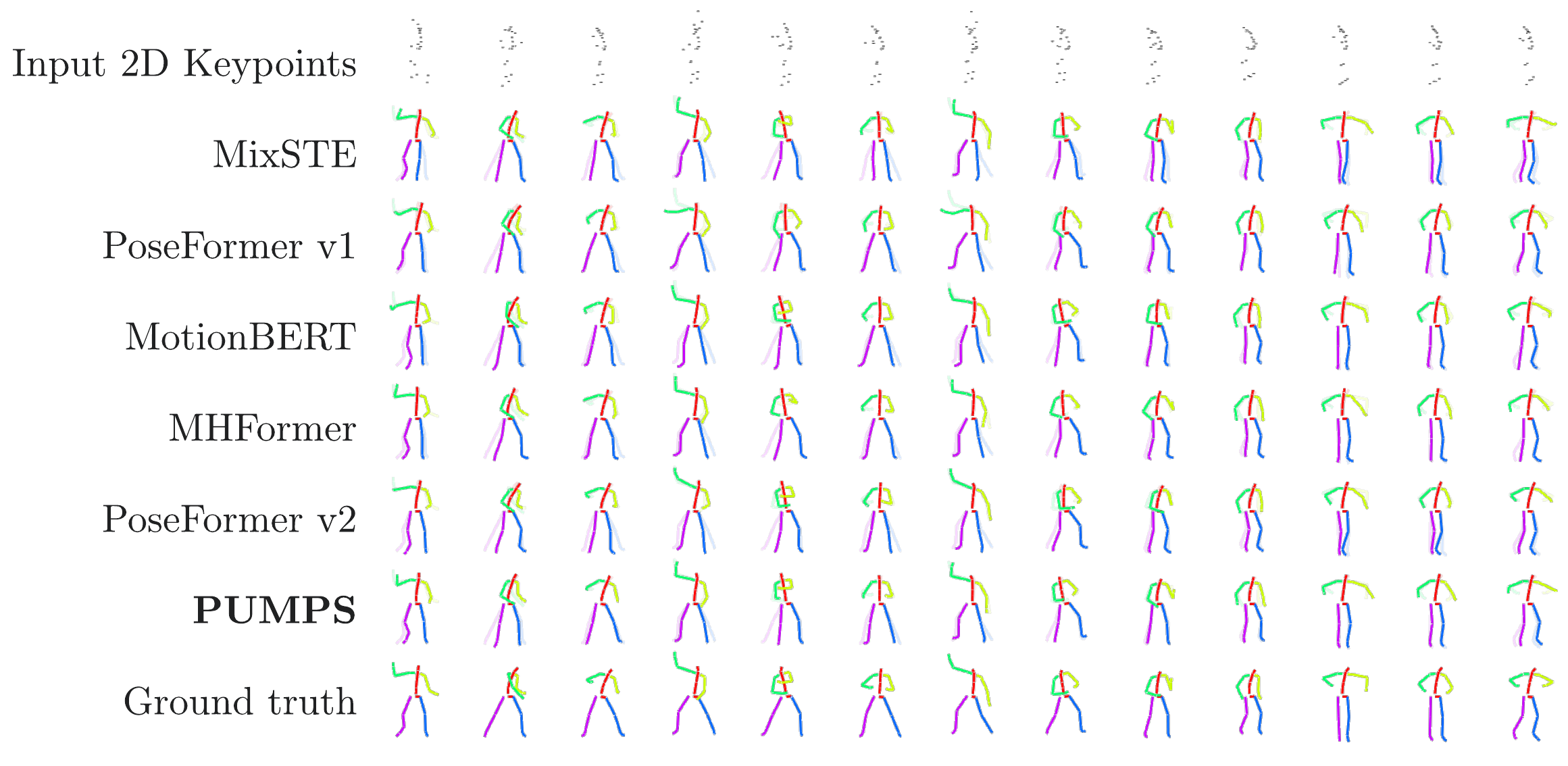}
    \caption{Motion estimation on a dance motion sample, given a sequence of 2D keypoints (top row). Predictions are rotated by $90\degree$ around the vertical axis for better visual comparison.}
    \label{fig:estimation}
\end{figure}

\subsection{Motion Denoising}

Motion denoising is crucial for regularising raw motion capture sequences, which are typically obtained  through  camera-based systems. 
Challenges during performance capture, including body part occlusion, camera synchronisation issues, and image noise, can cause defects into the captured motions. 
In response, motion denoising techniques aim to identify such defects and refine raw motions to ensure they adhere to realistic physical constraints and natural motion patterns. 
Like motion estimation, we employ joint-wise $\ell_2$ position and quaternion objectives. To generate noisy motion data for both training and evaluation, we introduce Gaussian noise into the pose space, with standard deviations of $\sigma \in (0, 0.1]$ for the 3D root joint position, and $\sigma \in (0, 0.2]$ for quaternion values. 
The noise is applied to between $20\%$ and $80\%$ of all frames per sequence.

Like with motion estimation, we compare two variants of PUMPS against several existing state-of-the-art methods. Our variants include a $\Phi^\text{LMS}$ denoising model trained from scratch, and a fine-tuned $\Phi^\text{LMS}$ denoising model. Motion denoising performance is measured in the aforementioned MPJPE for visual similarity, as well as per-joint acceleration error \citep{rempe2021humor} for motion smoothness. 
To compare with state-of-the-art methods, we conducted experiments with the VAE-based motion prior model 
HuMoR \citep{rempe2021humor}, the neural distance field optimisation method Pose-NDF \citep{tiwari2022posendf}, as well as end-to-end denoisers used in text-to-motion diffusion, specifically U-ViT \citep{bao2023all, chen2023executing} and the conventional Transformer model \citep{vaswani2017attention, liu2024human}. Furthermore, we provide results from Laplacian smoothing \citep{sorkine2004laplacian} as a classical reference point. For all learned methods, we use our LaFAN1 training split. 

\begin{table}
    \setlength\tabcolsep{3px}
    \begin{subtable}[h]{\linewidth}
        \resizebox{\linewidth}{!}{
            \begin{tabular}{l c c c c c c c c}
                \multirow{2}{*}{\textbf{MPJPE}$\downarrow$} & \multicolumn{8}{c}{\textit{Motion category}} \\
                \cmidrule(lr){2-9}
                 & \textit{Aim} & \textit{Dance} & \textit{Fall} & \textit{Fight} & \textit{Obst.} & \textit{Prone} & \textit{Push} & \textit{Walk} \\
                \cmidrule(lr){1-1} \cmidrule(lr){2-9}
                Laplacian & 61.27 & 69.80 & 64.40 & 69.89 & 63.39 & 62.14 & 64.55 & 63.97 \\
                Pose-NDF & 70.17 & 84.69 & 76.91 & 81.53 & 72.87 & 71.12 & 78.01 & 73.75 \\
                HuMoR & 75.74 & 107.06 & 97.93 & 105.68 & 76.51 & 80.91 & 93.38 & 76.73 \\
                U-ViT & 64.39 & 101.42 & 91.85 & 105.00 & 79.51 & 98.77 & 69.36 & 59.16 \\
                Transformer & 58.68 & 70.05 & 77.44 & 72.80 & 61.62 & 64.49 & 56.40 & 51.22 \\
                \cmidrule(lr){1-1}
                \textbf{PUMPS} (scratch) & 70.20 & 98.06 & 84.25 & 101.54 & 80.06 & 85.68 & 71.04 & 65.26 \\
                \textbf{PUMPS} (finetune) & 49.83 & 73.07 & 65.12 & 68.78 & 58.00 & 59.29 & 54.18 & 53.04 \\
                \bottomrule
            \end{tabular}
        }
    \end{subtable}
    \begin{subtable}[h]{\linewidth}
        \centering
        \resizebox{\linewidth}{!}{
            \begin{tabular}{l c c c c c c c c}
                \toprule
                \multirow{2}{*}{\textbf{Accel.}$\downarrow$} & \multicolumn{8}{c}{\textit{Motion category}} \\
                \cmidrule(lr){2-9}
                 & \textit{Aim} & \textit{Dance} & \textit{Fall} & \textit{Fight} & \textit{Obst.} & \textit{Prone} & \textit{Push} & \textit{Walk} \\
                \cmidrule(lr){1-1} \cmidrule(lr){2-9}
                Laplacian & 24.01 & 36.22 & 28.02 & 35.97 & 26.64 & 24.99 & 28.98 & 28.12 \\
                Pose-NDF & 54.93 & 68.26 & 60.21 & 66.83 & 57.54 & 55.94 & 60.37 & 58.38 \\
                HuMoR & 29.61 & 52.27 & 42.63 & 51.38 & 33.79 & 36.64 & 36.67 & 35.42 \\
                U-ViT & 19.10 & 44.97 & 28.85 & 41.21 & 25.97 & 28.99 & 26.69 & 22.22 \\
                Transformer & 18.54 & 33.95 & 24.29 & 32.70 & 22.65 & 22.23 & 23.73 & 20.42 \\
                \cmidrule(lr){1-1}
                \textbf{PUMPS} (scratch) & 8.92 & 33.05 & 15.89 & 31.41 & 14.46 & 11.99 & 17.90 & 14.79 \\
                \textbf{PUMPS} (finetune) & 10.45 & 32.27 & 16.55 & 30.51 & 15.31 & 12.76 & 18.87 & 16.33 \\
                \hline
            \end{tabular}
        }
    \end{subtable}
    \caption{Motion denoising method comparisons on 128-frame sequences, using MPJPE (top) and acceleration error (bottom).}
    \label{tab:denoise}
\end{table}
\begin{figure}
    \centering
    \includegraphics[width=\linewidth]{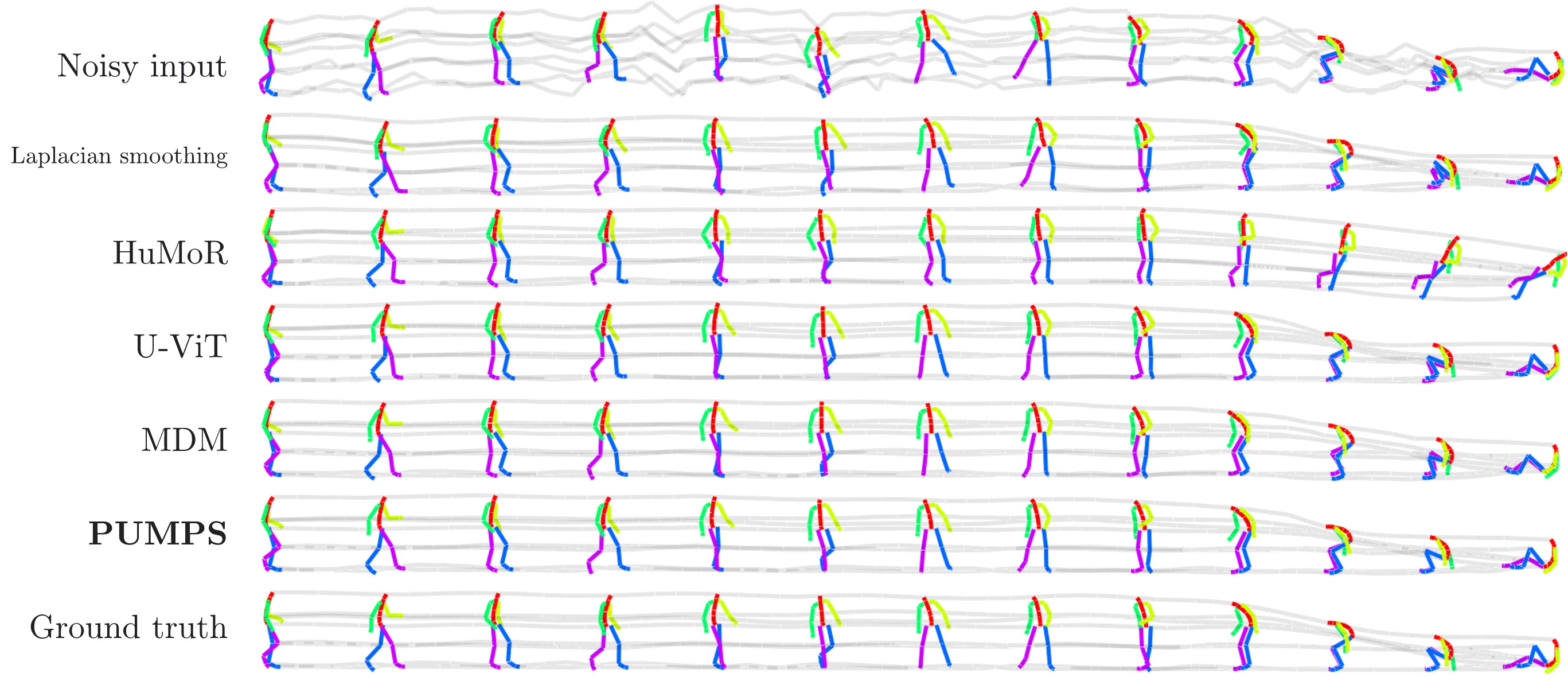}
    \caption{Motion denoising examples given a noisy falling motion sequence (top row). Though only every 5th frame is shown, all frames are included in the trajectory representation (grey lines).}
    \label{fig:denoise}
\end{figure}

The performance for each method is shown in Table \ref{tab:denoise}. Our fine-tuned method clearly demonstrates superior motion denoising performance compared to state-of-the-art approaches and Laplacian smoothing.
Variable motion priors appear to be ineffectual for MPJPE, as they typically end up measuring simple velocity and pose validity, rather than temporal motion naturalness features. End-to-end models, including our method, are stronger in this regard. In contrast, our approach's temporal attention mechanisms leverage pre-trained motion understanding to produce strong denoising results, significantly outperforming existing methods. As shown in Figure \ref{fig:denoise}, pre-trained motion knowledge allows PUMPS to recognise walking and falling motions, and generate a clean motion accordingly. 

Like with motion estimation, our method observes a notable improvement when fine-tuned from pre-trained weights, as opposed to training from scratch. This strongly suggests that the implicit motion patterns in the fine-tuned $\Phi^\text{LMS}$ model provides a crucial foundation for the synthesis component of motion denoising. Overall, these pre-trained features have led to a $\sim25\%$ decrease in MPJPE, albeit at a minor but acceptable cost to acceleration loss. 

%% file: sec/5_conclusion.tex
\section{Limitations and Future Work}
\label{sec:limitations}

Our method's ability to work in skeleton-specific pose space without supervision faces similar limitations as PC-MRL \citep{mo2024pointcloud}, the method we use to decode temporal point clouds into skeletal motion data. To re-iterate, the method requires either native or non-native motion datasets during training, and at least one known pose during evaluation due to its temporal offset-based evaluation. Furthermore, skeletal bones of undefined or zero length are incompatible with the method. In our experiments, all such bones were removed.


Modern diffusion-based motion synthesis models are primarily built on convolutional networks \citep{karunratanakul2023guided, zhang2024tedi}, and the applicability of attention-based architectures has been seldom explored in this emerging field. Since our method has shown strong efficacy in motion denoising processes, our future work entails explorations into motion diffusion applications. Additionally, as temporal point clouds are a relatively new medium for representing motion data, we are excited to expand to a broader range of motion-related tasks.

\section{Conclusion}
\label{sec:conclusion}

This paper primordially explores pre-training approaches for motion synthesis processes in the medium of skeleton-agnostic point clouds. 
The non-hierarchical nature of TPCs and its convertibility with skeletal data makes the representation medium ideal for producing unified motion synthesis models, which are preferred over skeleton-specific methods due to their universal applicability.
Our proposed method, PUMPS, employs a highly efficient latent sampling system using Gaussian noise vectors to enable an effective TPC auto-encoder architecture, which serves to train a foundational motion model in the TPC medium.
PUMPS strongly demonstrates superior performance for motion understanding and synthesis despite its non-native supervision strategy.  
